\documentclass[10pt,twocolumn,letterpaper]{article}

\usepackage{iccv}
\usepackage{times}
\usepackage{epsfig}
\usepackage{graphicx}
\usepackage{amsmath}
\usepackage{amssymb}

\usepackage[accsupp]{axessibility}

\usepackage{mathtools}
\usepackage{multirow}

\usepackage[breaklinks=true,bookmarks=false]{hyperref}

\iccvfinalcopy 

\def\iccvPaperID{****} 

\ificcvfinal\fi

\begin{document}


\title{Exploiting Spatial-Temporal Semantic Consistency for Video Scene Parsing}


\author{Xingjian He\textsuperscript{1,2}, Weining Wang\textsuperscript{1,2}, Zhiyong Xu\textsuperscript{1,3}, Hao Wang\textsuperscript{1,2}, Jie Jiang\textsuperscript{1,2},\\
Jing Liu\textsuperscript{1,2} \\
\textsuperscript{1} National Laboratory of Pattern Recognition, Institute of Automation, Chinese Academy of Sciences \\
\textsuperscript{2} School of Artificial Intelligence, University of Chinese Academy of Sciences \\
\textsuperscript{3} School of Automation and Electrical Engineering, University of Science and Technology Beijing \\
{\tt\small\texttt{\{xingjian.he,weining.wang,jie.jiang,jliu\}@nlpr.ia.ac.cn,}}\\
{\tt\small\texttt{g20198748@xs.ustb.edu.cn wanghao2019@ia.ac.cn}}}

\maketitle
\ificcvfinal\thispagestyle{empty}\fi

\begin{abstract}
Compared with image scene parsing, video scene parsing introduces temporal information, which can effectively improve the consistency and accuracy of prediction. In this paper, we propose a Spatial-Temporal Semantic Consistency method to capture class-exclusive context information. Specifically, we design a spatial-temporal consistency loss to constrain the semantic consistency in spatial and temporal dimensions. In addition, we adopt an pseudo-labeling strategy to enrich the training dataset. We obtain the scores of 59.84\% and 58.85\% mIoU on development (test part 1) and testing set of VSPW, respectively. And our method wins the 1st place on VSPW challenge at ICCV2021.

\end{abstract}

\section{Introduction}
Video scene parsing aims to assign pixel-wise semantic labels to each video frame. Directly using image scene parsing methods on video frames, without exploiting the temporal information of video, may lead to the problem of discontinuous prediction results.
Therefore, it is valuable to explore how to use video temporal information to improve the continuity of prediction.

\begin{figure}
    \centering
    \includegraphics[width=1\linewidth]{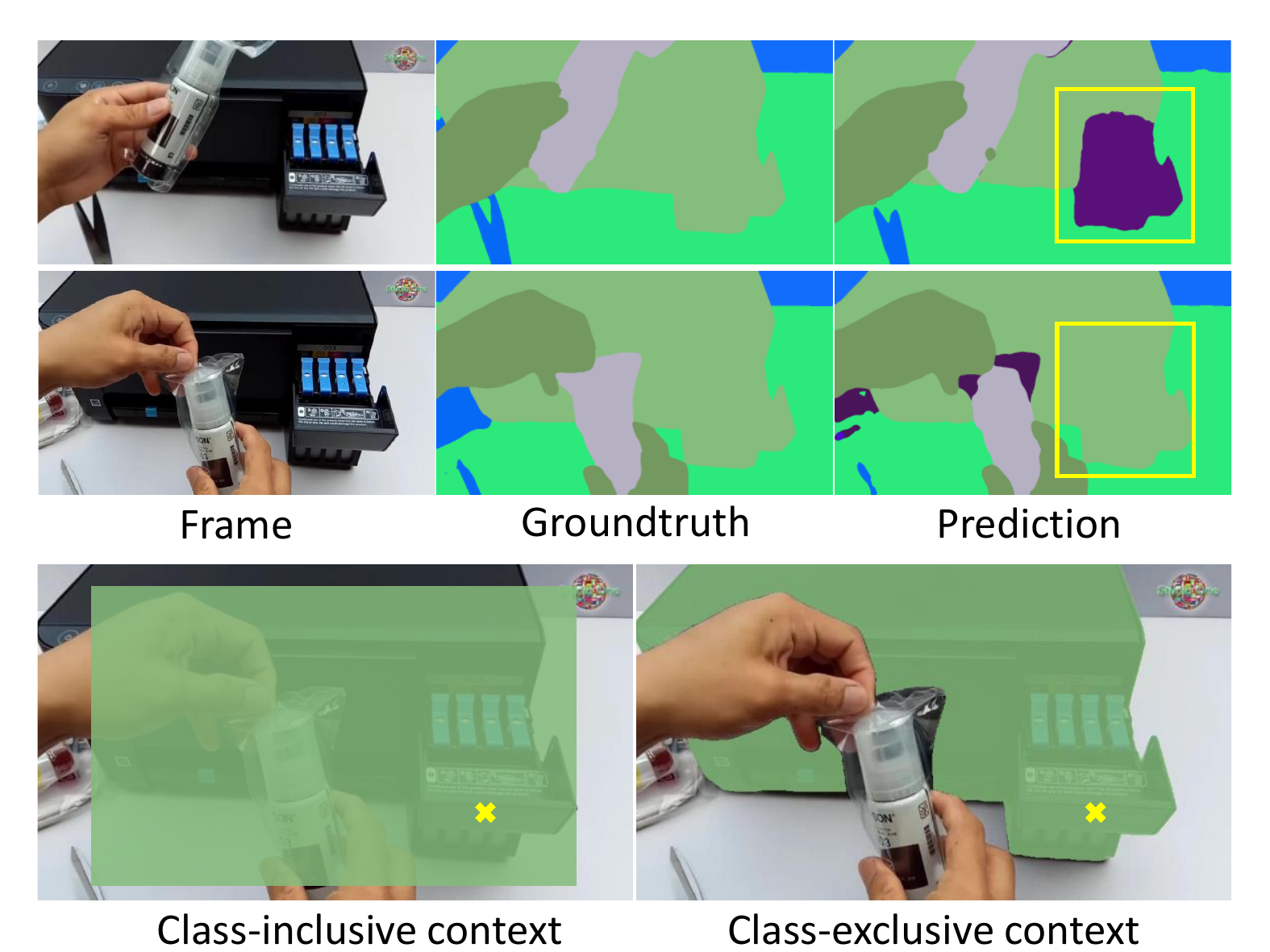}
    \caption{The first two rows show the discontinuity of predicted results on two adjacent frames. The third row shows the class-inclusive context which means information of pixels in all classes in the surroundings of the object and the class-exclusive context which means the information of pixels in the same category around the pixel.}
    \label{fig:one}
\end{figure}

By analyzing the prediction results of adjacent frames of multiple videos obtained by the image scene parsing, we find that when the object remains unchanged but the surrounding scene changes, the prediction results of the object may be inconsistent. As shown in Figure~\ref{fig:one}, in two adjacent frames, the ``printer" remains unchanged, while the surroundings of the ``printer" change. As a result, the predictions of the ``printer" in the two frames are inconsistent.
This is because these methods predict by learning the class-inclusive context, which represents information of pixels in all classes in the surroundings of the object. When the surroundings of the object changes, the prediction results of the object may also change, leading to discontinuous video prediction results.
Therefore, we consider that the model should learn the class-exclusive context information for prediction, in which the class-exclusive context represents the information of pixels in the same category around the pixel.
In this way, when the surrounding environment changes, the pixel prediction results between adjacent frames will remain relatively consistent.
The difference between the class-inclusive context and class-exclusive context is shown in Figure~\ref{fig:one} (bottom).

In this paper, we propose a Spatial-Temporal Semantic Consistency method to explore the class-exclusive context information.
Specifically, we take adjacent frames as the input of the network, and propose a spatial-temporal consistency loss (STCL) to constrain the intra-class consistency in spatial and temporal dimensions. 
Thus, the network can pay more attention to the feature of the object itself and reduce the dependence on other categories among the surrounding environment.
In other words, the model could capture class-exclusive context information and avoid noisy class-inclusive context information, thus the predicted results are more stable and robust. In addition, we put forward an pseudo labeling method to help the network improve the representation learning with limited labeled data.

We carry out extensive ablation experiments on Video Scene Parsing in the Wild (VSPW) dataset~\cite{miao2021vspw} to verify the effectiveness of our proposed method. We finally obtain the scores of 59.84\% and 58.85\% mIoU on test part 1 and testing set of VSPW, respectively. And our solution is ranked 1st place on VSPW challenge at ICCV2021.

\section{Method}

In this section, we first describe the overview pipeline of our network. And then, we introduce our Spatial-Temporal Semantic Consistency method. Next, we describe the pseudo-labeling strategy of unlabelled images. Finally, we describe the training and inference process of the network.

\begin{figure}
    \centering
    \includegraphics[width=1\linewidth]{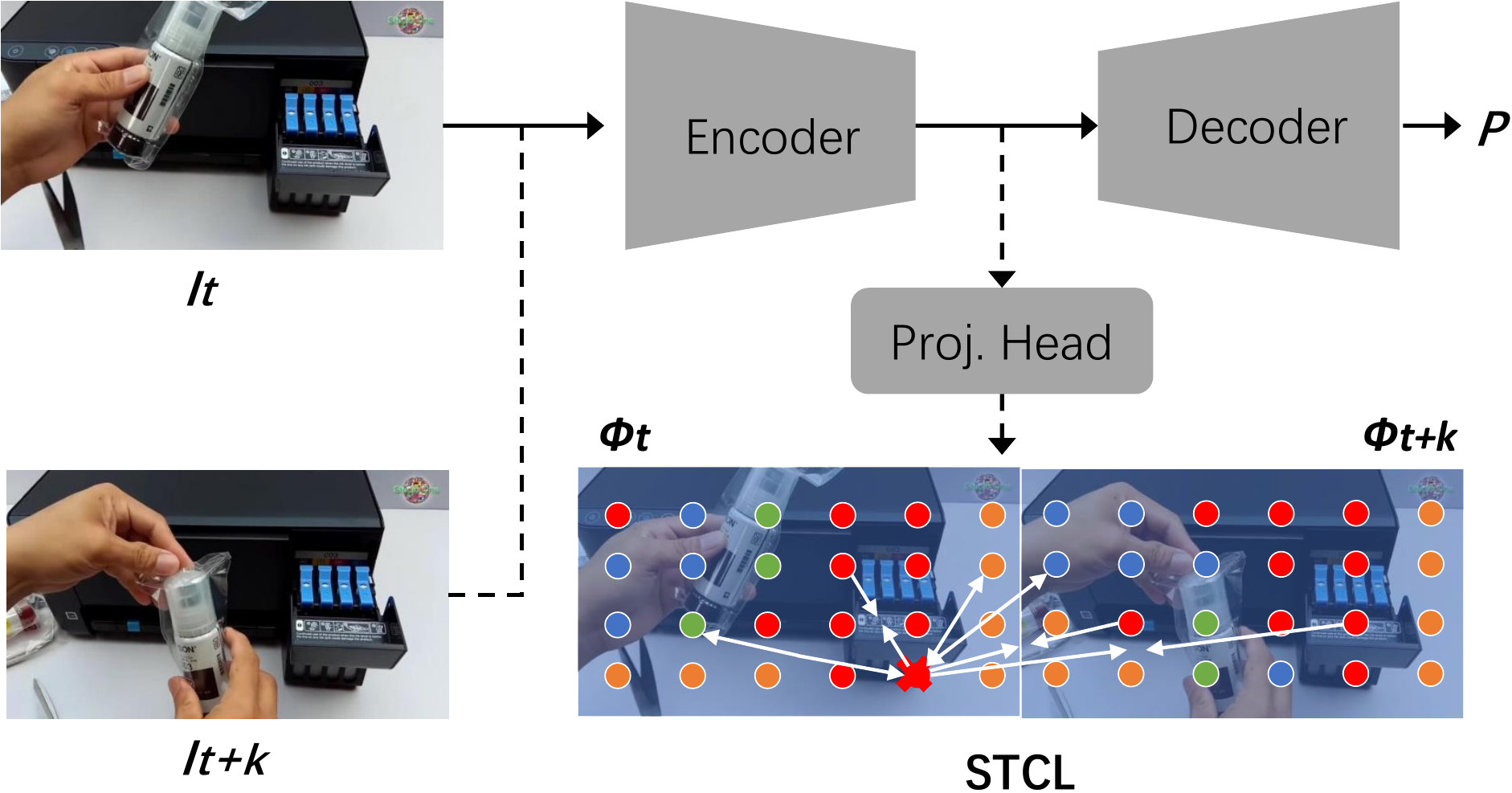}
    \caption{The overview pipeline of our network. STCL denotes spatial-temporal consistency loss.}
    \label{fig:two}
\end{figure}

\subsection{Overview}
As illustrated in Figure~\ref{fig:two}, given a video sequence, we take the current frame as the query frame, and a few adjacent frames of the current frame as the reference frames. 
The query and reference frames are fed into the encoder to extract the pixel-level feature. On the one hand, the encoded features pass through the decoder to recover the resolution for pixel-level prediction, which is supervised by the commonly used cross-entropy loss. On the other hand, the encoded features are transformed by the projection head, and the transformed features are supervised by the proposed spatial-temporal consistency loss.

{\flushleft{\bf{Encoder}}} We apply the CNN-based network (e.g. ResNet~\cite{he2016deep}, ResNeSt~\cite{zhang2020resnest}) and Transformer-based network (e.g. Swin Transformer~\cite{liu2021swin}) as encoder to extract image features.
{\flushleft{\bf{Decoder}}} Following UperNet~\cite{xiao2018unified}, we use Feature Pyramid Network~\cite{lin2017feature} to build high-level semantic feature maps at all scales and adopt multi-level feature aggregation for obtaining decoded features.

\subsection{Spatial-Temporal Semantic Consistency}

Most of the recent methods~\cite{fu2019dual, fu2020scene} construct class-inclusive context representation for each pixel to assist the semantic prediction. However, the class-inclusive context makes the network weaken the features learning of the object itself, resulting in the network heavily relying on class-inclusive context to achieve pixel prediction. For video sequences, the surroundings of the current pixel is fixed in spatial dimension, while may change in temporal dimension. When the surroundings changes and the object remains unchanged between the adjacent frames, the prediction of the object may be inconsistent. In this work, we propose a spatial-temporal consistency loss to capture class-exclusive context, establishing semantic consistency in spatial and temporal dimensions. The loss encourages the same category of features to be consistent, regardless of whether the surrounding environment is the same or not.

As shown in Figure~\ref{fig:two}, for input frames $I_t, I_{t+k}$ from the time $t$ and $t+k$, the $I_t, I_{t+k}$ first pass through the encoder to obtain the feature maps $F_t$ and $F_{t+k}$, respectively. Next, similar to~\cite{chen2020improved}, they are fed into a projection head to get transformed features $\phi_t$ and $\phi_{t+k}$, where the projection head consists of (FC)$\xrightarrow{}$(BN)$\xrightarrow{}$(ReLU). For a pixel $i$ with groundtruth semantic label $\hat{c}$, the positive samples are other pixels belonging to the same class $\hat{c}$ 
in $\phi_t$ and $\phi_{t+k}$, while the negatives are the pixels belonging to the other classes $C  \backslash  \hat{c}$. The $C$ denotes all classes in dataset. Formally, our spatial-temporal semantic consistency loss is defined as:

\begin{equation}\small
L(i) = \frac{1}{|P_i|}\sum_{i^+\in{P_i}}{-\log\frac{\exp(i\cdot i^+/\tau)}{\exp(i\cdot i^+/\tau)+\sum\limits_{i^-\in N_i}{\exp(i\cdot i^-/\tau)}}}
\label{eq::stl_loss_i}
\end{equation}

\begin{equation}\small
L_{stcl} = \frac{1}{|\Omega|}\sum_{i\in\Omega}L(i)
\label{eq::stl_loss}
\end{equation}
where $\Omega$ denotes the all pixels in $\phi_t$, $P_i$ and $N_i$ denote positive and negative samples for pixel $i$, respectively. $\tau$ denotes temperature coefficient.

By minimizing the above loss function, when  the  surrounding  environment  changes, the  pixel  prediction  results  between  adjacent  frames  will remain  relatively  consistent. Thus, the prediction results are more stable and accurate.

\subsection{Pseudo-Labeling Training}

it is usually partially labeled, such as vspw, which labels 15 f/s, and there are still a large number of unlabeled samples. Inspired by semi-supervised learning~\cite{chen2021semi,hung2018adversarial}, which aims to exploit unlabeled data to further improve the representation learning when given limited labeled data, we adopt an pseudo-labeling strategy for VSPW to enlarge the training dataset.

In particular, we first build a teacher network to provide a target probability for each of $N$ classes for every pixel in each image. And we adopt a hard labeling strategy, whereby for a given pixel, we select the top class prediction of the teacher network. In order to make pseudo label have higher confidence, we threshold the label based on teacher network output probability. Those pixels whose maximum prediction probability exceed the threshold will be set as true label, otherwise the pixels will be marked as ``ignore" class. In our experiments, the threshold is set to 0.5.

\subsection{Training \& Inference}
{\flushleft{\bf{Training}}} We jointly learn the semantic segmentation and the class-inclusive context information in an end-to-end mode. The total loss is defined as:

\begin{equation}
L = \lambda_1 L_{seg} + \lambda_2 L_{stcl}
\label{eq::total_loss}
\end{equation}
where $L_{seg}$ denotes cross-entropy loss, $\lambda_1$ and $\lambda_2$ are hyper-parameters used to balance the two losses. In practice, $\lambda_1=1$ and $\lambda_2=0.2$.

{\flushleft{\bf{Inference}}} In the inference phase, we process each frame of the input video independently. Therefore, no additional computational overhead is introduced in the inference stage.

\section{Experiments}

In this section, we first introduce the dataset and implementation details. After that we provide the ablation studies to validate the effectiveness of our proposed method. Finally, we report the results on the challenge test server. 

\subsection{Dataset and Evaluation Metrics}
The Video Scene Parsing in the Wild (VSPW) dataset~\cite{miao2021vspw} covers a  wide range of real-world scenarios and categories. Over 96\% of the captured videos are with high spatial resolutions from 720P to 4K. It also provides the dataset with the resolution of 480P. It densely annotates 3,536 videos, including 251,633 frames from 124 categories. The training set, validation set and testing set of VSPW contain 2,806/343/387 videos with 198,244/24,502/28,887 frames, respectively. 

In this paper, we adopt mean Interaction over Union (mIoU) and Weight IoU (WIoU) as evaluation metrics for scene parsing. In addition, to evaluate the stability across frames in a video, following~\cite{miao2021vspw}, we report Video Consistency (VC) metric to evaluate the category consistency among long-range adjacent frames. In particular, we report VC8 and VC16, which evaluate the consistency among 8 frames and 16 frames, respectively.

\subsection{Implementation details}

The Pytorch~\cite{paszke2017automatic} framework is employed to implement our network. In our experiments, we use CNN-based networks and Transformer-based networks as encoder. In training phase, for the CNN-based networks, the backbone is pretrained on ImageNet-1K~\cite{deng2009imagenet} and SGD optimizer is employed  with an initial learning rate of 0.01, a weight decay of 0.0001. For the Transformer-based networks, the backbone is pretrained on ImageNet-22K and AdamW optimizer is adopted with an initial learning rate of $ 7\times 10^{-5} $, and a linear warmup of 1,500 iterations. All models in our experiments are trained on 8 GPUs with 2 images per GPU for 20K iterations. For data augmentation, random horizontal flipping, random cropping (cropsize $768\times768$) and random resizing within ratio range [0.5, 2.0] are applied. During the testing phase, the sliding window method is used.

\subsection{Ablation Studies}

\begin{table}[t]
    \centering
    \begin{tabular}{lc|ccc}
    \hline
         {\bf Backbone} &{\bf Decoder} &{\bf mIoU} &{\bf VC8} &{\bf VC16}  \\
         \hline
         \hline
         ResNet-50  &UperNet   &0.3430	&0.8050	&0.7450	\\
         ResNest-101 &UperNet   &0.3758 &0.8430	&0.7920	 \\
         Cswin-L &UperNet   &0.5530	&0.8860	&0.8467	\\
         Swin-L  &UperNet   &\textbf{0.5627} &\textbf{0.8864} &\textbf{0.8539}	\\
         \hline
    \end{tabular}
    \caption{Experiments of different backbones on VSPW val set.  }
    \label{tab:backbone}
\end{table}

\begin{table}[t]
    \centering
    \begin{tabular}{l|ccc}
    \hline
         {\bf Method}  &{\bf mIoU} &{\bf VC8} &{\bf VC16}  \\
         \hline
         \hline
         Swin-L           &0.5627	&0.8864	&0.8539	\\
         Swin-L + ADE20K     &0.5670   &0.8792	&0.8451	 \\
         Swin-L + COCO    &0.5704	&0.8906	&0.8578	\\
         Swin-L + COCO-pre     &\textbf{0.5803}	&\textbf{0.8916} &\textbf{0.8611}	\\
         \hline
    \end{tabular}
    \caption{Experiments of extra data augmentation methods on VSPW val set.  }
    \label{tab:aug}
\end{table}

\begin{table}[t]
    \small
    \centering
    \begin{tabular}{l|ccccc}
    \hline
         {\bf Method} &{\bf dim} &{\bf $\tau$} &{\bf mIoU} &{\bf VC8} &{\bf VC16}  \\
         \hline
         \hline
         Swin-L        &-   &-              &0.5803	&0.8916	&0.8611	\\
         Swin-L + STCL  &512 &0.07            &\textbf{0.5930} &\textbf{0.9007} &\textbf{0.8687} \\
         Swin-L + STCL  &512 &0.05            &0.5876 &0.8989 &0.8649 \\
         Swin-L + STCL  &256 &0.07            &0.5872 &0.8986 &0.8651 \\
         \hline
    \end{tabular}
    \caption{Experiments of spatial-temporal consistency loss on VSPW val set.}
    \label{tab:st}
\end{table}

\begin{table}[!h]
    \centering
    \small
    \begin{tabular}{l|c|ccc}
    \hline
         \multirow{2}*{\textbf{Method}}  &{val set}  &\multicolumn{3}{c}{Test part 1}  \\ 
         \cline{2-5}
                       &{\bf mIoU}  &{\bf mIoU} &{\bf VC8} &{\bf VC16}  \\
         \hline
         \hline
         Swin-L + STCL          &0.5930  &-	&-	&-	\\
         Teacher          &\bf{0.6085}  &0.5638	&0.9120	&0.8769	 \\
         Swin-L + STCL + Ps.  &0.6006 &\bf{0.5743} &\bf{0.9253} &\bf{0.8989}	\\

         \hline
    \end{tabular}
    \caption{Experiments of pseudo labeling methods on VSPW val set and development (test part 1). Ps. denotes Pseudo labeled data.}
    \label{tab:pesudo_label}
\end{table}

\begin{table}[]
    \centering

    \begin{tabular}{l|cccc}
    \hline
         {\bf Team}    &{\bf mIoU}  &{\bf WIoU}  &{\bf VC8}       &{\bf VC16}  \\
         \hline
         \hline
         CASIA\_IVA          &\bf{0.5984} &\bf{0.7307} &\bf{0.9451} &\bf{0.9222}\\
         BetterThing         &0.5806   &0.7205       &0.9325	&0.9053	 \\
         CharlesBLWX         &0.5738   &0.7210	     &0.9046	&0.8669	\\
         ustcvim             &0.5497   &0.7188	     &0.9017	&0.8674	\\
         Arlen               &0.5477   &0.6991       &0.9127    &0.8825 \\
         \hline
    \end{tabular}
    \caption{Comparisons with other methods on the VSPW-challenge-2021 (development) test part 1.}
    \label{tab:dev}
\end{table}

\begin{table}[]
    \centering

    \begin{tabular}{l|cccc}
    \hline
         {\bf Team}    &{\bf mIoU}  &{\bf WIoU}  &{\bf VC8}       &{\bf VC16}  \\
         \hline
         \hline
         CASIA\_IVA          &\bf{0.5885} &\bf{0.7178} &\bf{0.9477} &\bf{0.9259}\\
         CharlesBLWX         &0.5744   &0.7205       &0.9129	&0.8770	 \\
         BetterThing         &0.5735   &0.7210	     &0.9328	&0.8621	\\
         Arlen               &0.5562   &0.6963       &0.8997    &0.8621 \\
         ustcvim             &0.5549   &0.7038	     &0.9094	&0.8746	\\
         \hline
    \end{tabular}
    \caption{Comparisons with other methods on the VSPW-challenge-2021 final test set.}
    \label{tab:final}
\end{table}

{\flushleft{\bf{Backbone}}}
We explore the CNN-based and Transformer-based networks as our backbone. The CNN-based networks employ convolutional operations to extract semantic information, which are often affected by limited receptive fields in recognition. The Transformer-based networks employ the self-attention mechanism to model long-range dependencies and obtain competitive results. With large dataset for pre-training, the Transformer-based models show a strong ability of feature representation. The experimental results of different backbones are shown in Table~\ref{tab:backbone}. From the table we can find that, the Transformer-based networks (Cswin-L~\cite{dong2021cswin}, Swin-L~\cite{liu2021swin}) outperform the CNN-based networks (ResNet~\cite{he2016deep}, ResNeSt~\cite{zhang2020resnest}) for a large margin, which demonstrates that the Transformer-based networks are strong to extract image features. In the following experiments, we employ Swin-L as our backbone network.

{\flushleft{\bf{Extra Data Augmentation}}}
To enrich the dataset, we also utilize extra data from other datasets for training and pre-training. ADE20K dataset~\cite{zhou2017scene} is a dataset for image scene parsing task that spans diverse annotations of scenes, objects, parts of objects, and in some cases even parts of parts. It contains nearly 20K densely annotated images and 150 classes, which can provide a lot of additional training data. COCO dataset~\cite{lin2014microsoft} yields approximately 118K images in total and we use the images with panoptic labels for training in experiments. As shown in Table~\ref{tab:aug}, when we add ADE20K dataset and COCO dataset in the training phase, the model obtains improvements with these extra data. Further, when employing COCO dataset in the pre-training phase, the model achieves greater improvements (from 0.5627 to 0.5803), which demonstrates that extra data is more suitable for pre-training.

{\flushleft{\bf{Spatial-Temporal Consistency Loss}}}
We propose a spatial-temporal consistency loss to improve the video consistency and accuracy of prediction results. The experimental results are shown in Table~\ref{tab:st}. We analysis the model performance under different feature dimensions and temperature coefficients. The results show that with dimension set to 512 and $\tau$ set to 0.07, the model improves the video consistency (VC8) of the basic model from 0.8916 to 0.9007, demonstrating the effectiveness of our method.

{\flushleft{\bf{Pseudo-Labeling Training}}}
To further make full use of the unlabelled data, we apply an pseudo-labeling strategy to generate pseudo labels on unlabelled data. We combine the pseudo data and training set to obtain a new dataset. Then, we train our model on the new dataset to improve the performance. As shown in Table \ref{tab:pesudo_label}, we first build a teacher network, which ensembles Swin-L+ADE20K (in Table \ref{tab:aug}) and Swin-L+STCL (in Table \ref{tab:st}) models. The teacher model achieves 0.6085 on val set. And we submit the prediction results on testing set to competition server, obtaining 0.5670, 0.8792 and 0.8451 in terms of mIoU, VC8 and VC16, respectively. We use the strong teacher model to generate pseudo labels and train the model with the obtained pseudo labels. We achieve 0.5704, 0.8906 and 0.8578 in terms of mIoU, VC8 and VC16, respectively.

\subsection{Comparisons}

We use different image ratios of [0.5, 0.75, 1.0, 1.25, 1.5, 1.75] for multi-scale testing, and apply horizontal flipping for each scale. Moreover, we ensemble three different models for boosting the performance. Table \ref{tab:dev} and Table \ref{tab:final} show that our method surpasses others by a large margin. Specifically, in the development (test part 1), our solution achieves 0.5984 mIoU, 0.7307 WIoU, 0.9451 VC8, 0.9222 VC16, which is a large gap to the second method on four evaluation metrics. In the final testing set, our model maintains the 1st place on the four metrics.

\section{Conclusion}
In this work, we propose a Spatial-Temporal Semantic Consistency method for video scene parsing. We propose a spatial-temporal consistency loss to encourage the model to learn class-exclusive context information, improving intra-class consistent. In addition, we employ an pseudo-labeling strategy to make full use of unlabelled data to enrich the training dataset. With the proposed method and other practical tricks, e.g. stronger backbone, extra data augmentation, and ensembling method, our solution outperforms the others significantly and wins the 1st place on VSPW challenge at ICCV2021.

\section*{Acknowledgements}
This work was supported by the National Key Research and Development Program of China (No. 2020AAA0106400), National Natural Science Foundation of China (61922086, 61872366), and Beijing Natural Science Foundation (4192059, JQ20022).

{\small
\bibliographystyle{ieee_fullname}
\bibliography{egbib}

\begin{thebibliography}{10}\itemsep=-1pt

\bibitem{chen2020improved}
Xinlei Chen, Haoqi Fan, Ross Girshick, and Kaiming He.
\newblock Improved baselines with momentum contrastive learning.
\newblock {\em arXiv preprint arXiv:2003.04297}, 2020.

\bibitem{chen2021semi}
Xiaokang Chen, Yuhui Yuan, Gang Zeng, and Jingdong Wang.
\newblock Semi-supervised semantic segmentation with cross pseudo supervision.
\newblock In {\em Proceedings of the IEEE Conference on Computer Vision and
  Pattern Recognition}, pages 2613--2622, 2021.

\bibitem{deng2009imagenet}
Jia Deng, Wei Dong, Richard Socher, Li-Jia Li, Kai Li, and Li Fei-Fei.
\newblock Imagenet: A large-scale hierarchical image database.
\newblock In {\em Proceedings of the IEEE Conference on Computer Vision and
  Pattern Recognition}, pages 248--255. Ieee, 2009.

\bibitem{dong2021cswin}
Xiaoyi Dong, Jianmin Bao, Dongdong Chen, Weiming Zhang, Nenghai Yu, Lu Yuan,
  Dong Chen, and Baining Guo.
\newblock Cswin transformer: A general vision transformer backbone with
  cross-shaped windows.
\newblock {\em arXiv preprint arXiv:2107.00652}, 2021.

\bibitem{fu2020scene}
Jun Fu, Jing Liu, Jie Jiang, Yong Li, Yongjun Bao, and Hanqing Lu.
\newblock Scene segmentation with dual relation-aware attention network.
\newblock {\em IEEE Transactions on Neural Networks and Learning Systems},
  2020.

\bibitem{fu2019dual}
Jun Fu, Jing Liu, Haijie Tian, Yong Li, Yongjun Bao, Zhiwei Fang, and Hanqing
  Lu.
\newblock Dual attention network for scene segmentation.
\newblock In {\em Proceedings of the IEEE Conference on Computer Vision and
  Pattern Recognition}, pages 3146--3154, 2019.

\bibitem{he2016deep}
Kaiming He, Xiangyu Zhang, Shaoqing Ren, and Jian Sun.
\newblock Deep residual learning for image recognition.
\newblock In {\em Proceedings of the IEEE Conference on Computer Vision and
  Pattern Recognition}, pages 770--778, 2016.

\bibitem{hung2018adversarial}
Wei-Chih Hung, Yi-Hsuan Tsai, Yan-Ting Liou, Yen-Yu Lin, and Ming-Hsuan Yang.
\newblock Adversarial learning for semi-supervised semantic segmentation.
\newblock {\em arXiv preprint arXiv:1802.07934}, 2018.

\bibitem{lin2017feature}
Tsung-Yi Lin, Piotr Doll{\'a}r, Ross Girshick, Kaiming He, Bharath Hariharan,
  and Serge Belongie.
\newblock Feature pyramid networks for object detection.
\newblock In {\em Proceedings of the IEEE Conference on Computer Vision and
  Pattern Recognition}, pages 2117--2125, 2017.

\bibitem{lin2014microsoft}
Tsung-Yi Lin, Michael Maire, Serge Belongie, James Hays, Pietro Perona, Deva
  Ramanan, Piotr Doll{\'a}r, and C~Lawrence Zitnick.
\newblock Microsoft coco: Common objects in context.
\newblock In {\em Proceedings of the European Conference on Computer Vision},
  pages 740--755. Springer, 2014.

\bibitem{liu2021swin}
Ze Liu, Yutong Lin, Yue Cao, Han Hu, Yixuan Wei, Zheng Zhang, Stephen Lin, and
  Baining Guo.
\newblock Swin transformer: Hierarchical vision transformer using shifted
  windows.
\newblock {\em arXiv preprint arXiv:2103.14030}, 2021.

\bibitem{miao2021vspw}
Jiaxu Miao, Yunchao Wei, Yu Wu, Chen Liang, Guangrui Li, and Yi Yang.
\newblock Vspw: A large-scale dataset for video scene parsing in the wild.
\newblock In {\em Proceedings of the IEEE Conference on Computer Vision and
  Pattern Recognition}, 2021.

\bibitem{paszke2017automatic}
Adam Paszke, Sam Gross, Soumith Chintala, Gregory Chanan, Edward Yang, Zachary
  DeVito, Zeming Lin, Alban Desmaison, Luca Antiga, and Adam Lerer.
\newblock Automatic differentiation in pytorch.
\newblock 2017.

\bibitem{xiao2018unified}
Tete Xiao, Yingcheng Liu, Bolei Zhou, Yuning Jiang, and Jian Sun.
\newblock Unified perceptual parsing for scene understanding.
\newblock In {\em Proceedings of the European Conference on Computer Vision},
  pages 418--434, 2018.

\bibitem{zhang2020resnest}
Hang Zhang, Chongruo Wu, Zhongyue Zhang, Yi Zhu, Haibin Lin, Zhi Zhang, Yue
  Sun, Tong He, Jonas Mueller, R Manmatha, et~al.
\newblock Resnest: Split-attention networks.
\newblock {\em arXiv preprint arXiv:2004.08955}, 2020.

\bibitem{zhou2017scene}
Bolei Zhou, Hang Zhao, Xavier Puig, Sanja Fidler, Adela Barriuso, and Antonio
  Torralba.
\newblock Scene parsing through ade20k dataset.
\newblock In {\em Proceedings of the IEEE Conference on Computer Vision and
  Pattern Recognition}, pages 633--641, 2017.

\end{thebibliography}
}

\end{document}